\documentclass[
	a4paper, 
	10pt, 
	unnumberedsections, 
	twoside, 
]{t0004}
\usepackage{float}
\usepackage[title]{appendix}
\usepackage{amsfonts}
\usepackage{xcolor}

\runninghead{Action Recognition YGAR} 
\title{Action Recognition\\ Utilizing YGAR Dataset} 
\author{
	Shuo Wang, Amiya Ranjan and Lawrence Jiang
}
\date{\footnotesize DATASCI 281 Computer Vision \\ UC Berkeley School of Information \\ \{shuo.wang2, aranjan, lawrencejiang0797\}@berkeley.edu}
\renewcommand{\maketitlehookd}{%
	\begin{abstract}
		\noindent The scarcity of high quality actions video data is a bottleneck in the research and application of action recognition. Although significant effort has been made in this area, there still exist gaps in the range of available data types a more flexible and comprehensive data set could help bridge. In this paper, we present a new 3D actions data simulation engine and generate 3 sets of sample data to demonstrate its current functionalities. With the new data generation process, we demonstrate its applications to image classifications, action recognitions and potential to evolve into a system that would allow the exploration of much more complex action recognition tasks. In order to show off these capabilities, we also train and test a list of commonly used models for image recognition to demonstrate the potential applications and capabilities of the data sets and their generation process.
	\end{abstract}
}

\begin{document}
\maketitle 

\section{Introduction}

Action recognition is an important area of research within the field of machine learning. Potential applications of movement recognition and understanding of human actions are endless, ranging from robotics and security surveillance to human-machine interactions. Successful progress in this field would translate into solutions for many real-world problems. Not only are the potentials enormous, diverse and distinct research topics exist within the broad category of action recognition. Single action classification\cite{Carreira:2018qr} is one of the more widely studied topics, and shares many similar techniques to image classification \cite{Ren:2015qr}. Moving beyond single actions, localized multiple object action recognition are also possible\cite{Wu:2023qr, Gu:2018qr}. Even sequential action recognition has been attempted\cite{Yeung:2017qr}. Although tremendous progress has been made, much still remains to be done. Some of the obstacles facing the research efforts include: relative scarcity of high quality data\cite{Carreira:2018qr}, high resource requirements for training potential model architectures and difficulty in the processing and modifications of input data for analysis.

In this paper, we propose a new method of video actions data generation by means of 3D simulation, where data generation could be customized to facilitate various research topics and specific areas of focus, including single action recognition, action orientation detection and action segmentation. In addition, we also perform several tests of action recognition using classic image classification modeling techniques and deep learning techniques, demonstrating how our dataset could be leveraged to bridge the gap between image classification tasks and 3D action recognition. Finally we test the effecacies of video vision transformers \cite{Aarnab:2021qr} on our data sets to compare with the results from image classifiers.

\section{Background}

\subsubsection{Dataset} Several data sets exist currently for single action recognition, including UCF101\cite{Soomro:2012qr} and HMDB\cite{Kuehne11}. UCF101 dataset is a collection of youtube videos that contains 101 action categories with a total of 13320 videos, averaging 132 videos per category. Some of the categories in the dataset include ``Apply Eye Makeup'', ``archery'' and ``Frisbee catch''.

HMDB\cite{Kuehne11} is another data set that aims to further the progress in understanding action recognition. It contains 51 actions and a total of 7000 video clips. The actions within the videos are broadly categorized into five types: general facial actions, facial actions with object manipulation, general body movements, body movements with object interaction and body movements for human interaction.

Over time, larger data sets have been created for video action recognition tasks such as the kinetics dataset\cite{Smaira:2020qr}, where 700 classes of actions have been categorized, each with an average of 926 sample videos.

More recently, spatio-temporally localized atomic visual actions data sets (AVA)\cite{Gu:2018qr} have also been introduced, where actions by multiple objects exist within the same video sample.

\subsubsection{Action Recognition} As more and more data sets come into existence, models have been built to train and test on these data sets, including convolutional neural network, LSTM\cite{Carreira:2018qr} based encoder models and 3D neural networks that combines frames and optical flow information \cite{Carreira:2018qr}. More complex architectures for localized action recognitions have also been proposed\cite{Wu:2023qr}.

\subsubsection{Bottlenecks} Although commendable progress has been made in the research of action understanding, many issues still hinder its progress. One of them is the quality of available data sets. The existing data sets are typically collected from video data sources and curated by human judges, this process introduces many variabilities in the quality and comprehensiveness of the sample data, and because these data are collected as they are available, often it is impossible to control for characteristics that are desired for particular research objectiveness. For example, it is often difficult to study the specific effects of object variation and orientation separately. Besides targeted research needs, it is often difficult to study hierarchies of categories. As an example, if we would like to conduct a research where we would like to first recognize the person in the image, and then the action performed by the person, then it would be very difficult to conduct such studies with the existing data sets.

\begin{figure}
	\includegraphics[width=\linewidth]{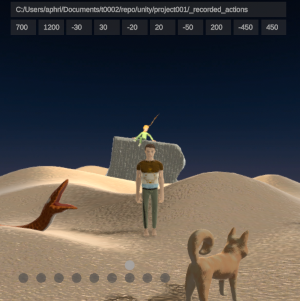}
	\caption{Action simulation engine.}
	\label{fig:simulationengine}
\end{figure}

The second issue is the quantity of available data, although kinetics data set has attempted to address the issue of volume with larger numbers of samples, they are still only single action samples of variable qualities, making studies of localized actions and sequential action recognition difficult.

We hope that our new data set could help address some of these issues and contribute to the progress of the study in action recognition.

\section{Data}

\subsection{Data Generation}

\begin{figure}
	\includegraphics[width=\linewidth]{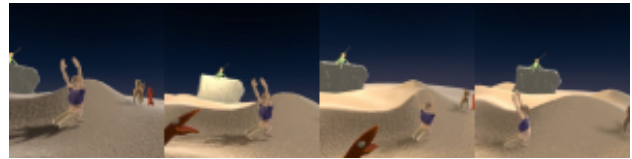}
	\caption{Various camera angle, offset and zoom configurations.}
	\label{fig:orientationfig}
\end{figure}

Our data set is generated by a 3D simulation program(Figure \ref{fig:simulationengine}) developed in Unity that supports configurations for zoom, center offset, camera angle orientation and avatar styles. The amount of zoom applied to the camera typically varies from 50\% to 200\%, 100\% represents the default zoom amount. An offset could be applied to the x or y direction of the camera relative to the target of the camera, the unit of this configuration is based on the simulation world space metric system (meters in world space). The camera could also be rotated about the x and y axis relative to the target avatar; we typically set these configurations between -5 degrees to 90 degrees about the x axis and -90 degrees to 90 degrees about the y axis(Figure \ref{fig:orientationfig}).

\begin{figure}
	\includegraphics[width=\linewidth]{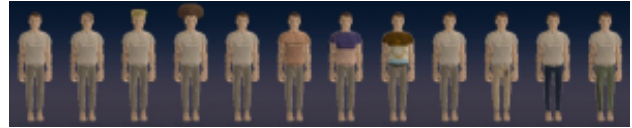}
	\caption{Appearance styles. First set of 4 figures show possible hairstyles. Second set of 4 figures show possible cloth patterns. Third set of 4 figures shows possible pants patterns.}
	\label{fig:appearancefig}
\end{figure}

The avatar character in the simulation could be configured with various hair, cloth and pants styles. In our current iteration, there are 4 different styles for hair, cloth and pants respectively, for a total of 64 unique combinations(Figure \ref{fig:appearancefig}).

Our first set of actions include a total of 10 different yoga poses: camel, chair, childs, lord of the dance, lotus, thunderbolt, triangle, upward dog, warrior II and warrior III. Each of the 10 yoga poses have 4 type variations within them, some with more pronounced difference than others, for a total of 40 action and action types(Figure \ref{fig:actionsfig}). We have decided to choose this set of actions as our first action sets because each pose is distinct and has a defined ending position, which allows us to use image classification techniques to compare classic modeling architecture and deep learning ones.

Our simulation also supports the option to include static background and dynamic background objects which would allow us to adjust the complexity of the sample data.

\subsection{Data Set}

From the actions and configurations available within our simulation program, we generated 3 sets of video actions data of varying difficulties based on the zoom, offset, angle and scene background configurations specified: easy, medium and hard.

\begin{figure}
	\includegraphics[width=\linewidth]{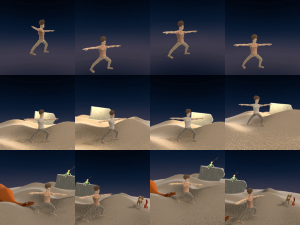}
	\caption{Difficulty levels. First row shows samples from easy level data set. Second row shows samples from medium level data set. Third row shows samples from hard level data set.}
	\label{fig:difficultylevels}
\end{figure}

The configurations of the three datasets are listed in the Table \ref{tab:dataconfig}. Each dataset is generated as follows: for each of the 40 action and action types, we sample 25 random hair, cloth and pants style combinations to create action scenes for, and for each action scene, we capture the action with 20 cameras of randomly generated offsets and angles based on the constraints specified. Each dataset is therefore consisted of 20,000 videos, every action and action type label combination contains 500 videos and every action label contains 2000 videos. Figure \ref{fig:difficultylevels} shows samples from each difficulty level.

\begin{table} 
	\caption{Data set configurations: zoom in percentage, offset in meters, angle in degrees.}
	\centering
	\begin{tabular}{lccc}
		\toprule
		Type & Easy & Medium & Hard \\
		\midrule
		Min. Zoom & 80\% & 80\% & 70\% \\
		Max. Zoom & 110\% & 110\% & 120\% \\
		\hline
		Min. X Offset & -1.5 & -1.5 & -3.0 \\
		Max. X Offset & 1.5 & 1.5 & 3.0 \\
		\hline
		Min. Y Offset & -1.5 & -1.5 & -2.0 \\
		Max. Y Offset & 1.5 & 1.5 & 2.0 \\
		\hline
		Min. X Angle & -5$^{\circ}$ & -5$^{\circ}$ & -5$^{\circ}$ \\
		Max. X Angle & 10$^{\circ}$ & 10$^{\circ}$ & 20$^{\circ}$ \\
		\hline
		Min. Y Angle & -30$^{\circ}$ & -30$^{\circ}$ & -45$^{\circ}$ \\
		Max. Y Angle & 30$^{\circ}$ & 30$^{\circ}$ & 45$^{\circ}$ \\
		\hline
		Static Background & Off & On & On \\
		Dynamic Background & Off & Off & On \\
		\bottomrule
	\end{tabular}
	\label{tab:dataconfig}
\end{table}

As we can see from Table \ref{tab:dataconfig}, the configuration of the easy and medium dataset is identical except the medium dataset has the static background turned on.

Each video is typically about 1 second in length, sampled at 30 FPS. The size of each video is around 80KB and the frame is 351X351 pixels.

\begin{figure}
	\includegraphics[width=\linewidth]{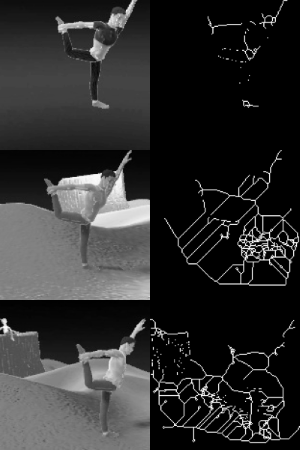}
	\caption{Skeletonizing of images. First row shows easy level sample, second row shows medium level sample and third row shows hard level sample.}
	\label{fig:sklimages}
\end{figure}

\section{Model}

Before creating models for the data sets, we have extracted the frame at 50\% location of each video as input to our models, since the actions within each video are yoga poses and could be adequately represented by the middle frame. This approach would ideally be enhanced later to use multiple frames within each video, supplemented with temporal information represented by optical flow.

Each frame was extracted from the video as 256X256X3 pixel RBG image, then converted to grayscale image of 256X256 pixels. Afterwards, we cropped the top, bottom, left and right of each grayscale image by 70, 30, 50 and 50 pixels to create 156X156 pixels grayscale image, in order to further reduce the size of input data.

\subsection{Filter}

\subsubsection{Principal Component Analysis} For computing PCA, we first downsized the input grayscale frame by 50\% to 78X78 pixels and flatten the image into an array of length 6084. Then we applied PCA on the input array, keeping the top 256 principal components. Finally each of the image arrays was projected onto the top 256 components to create the PCA weights for each image.

\subsubsection{Skeletonize} In order to construct image skeletons, we use the thinning algorithm proposed in \cite{TYZhang:1984qr}, which uses two iterations of pixel deletion to reduce the image down to a ``skeleton'' of unitary thickness. Figure \ref{fig:sklimages} shows the before and after results of applying the skeletonize algorithm on the ``easy'' level data set.

\subsubsection{Histogram of Oriented Gradients} 
\begin{figure}
	\includegraphics[width=\linewidth]{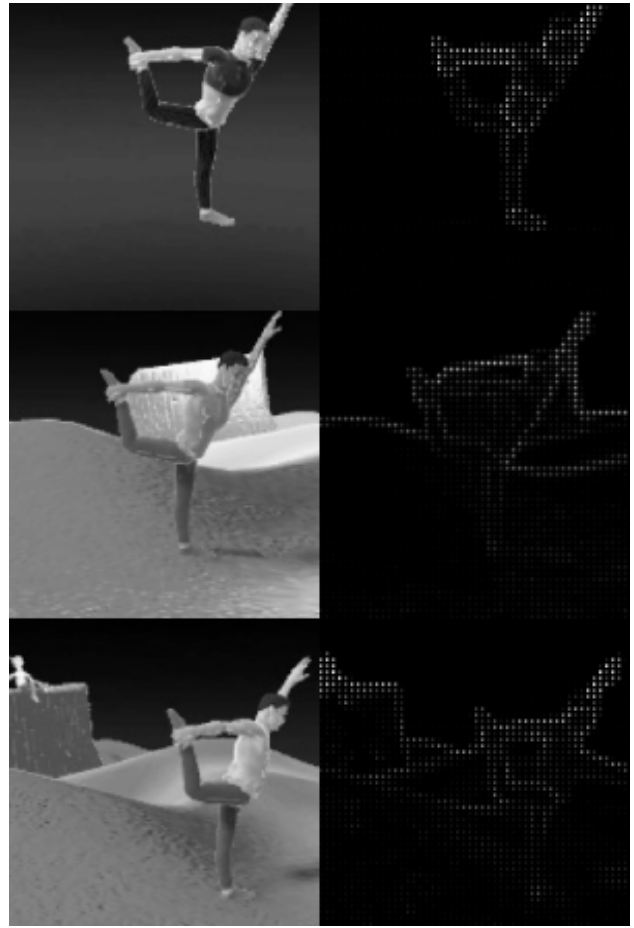}
	\caption{HOG filtering of images. First row shows easy level sample, second row shows medium level sample and third row shows hard level sample.}
	\label{fig:hogimages}
\end{figure}

In order to compute the HOG features, we first cropped the input grayscale image to 100X100 pixels, due to the amount of time HOG computation required. Then the HOG feature descriptor was computed with 9 orientations, 2x2 pixels per cell and 2x2 cells per block. Figure \ref{fig:hogimages} shows the results of the filter applied to images of different difficulty levels. From the sample image, it's clear that HOG filtered images are best for images without much structure in the background. As the complexity of the background increases, the HOG features of the avatar become less distinguishable .

\subsubsection{Scale-Invariant feature transform \& K-Mean Cluster} 
\begin{figure}
	\includegraphics[width=\linewidth]{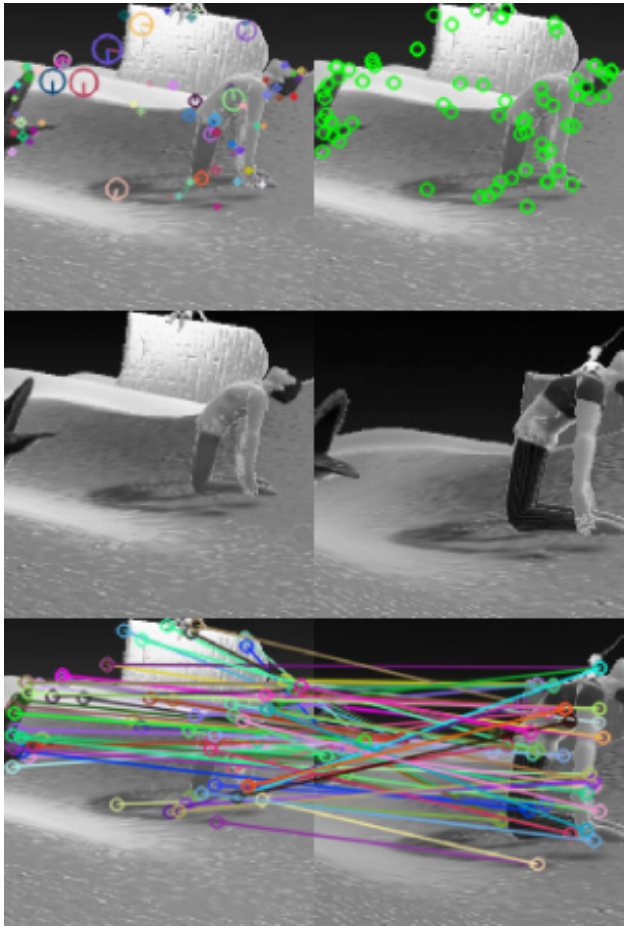}
	\caption{SIFT feature mapping. First row shows sift features (with and without size) extracted from train image. Second row shows the train image and test image in gray scale. Third row shows the train image features matched to test image features.}
	\label{fig:siftres}
\end{figure}

Bag of words is a commonly used technique in image classification. Similar to NLP, image features are used as words. We utilized SIFT features for this purpose. SIFT is one of the important algorithms that detect objects irrelevant to the scale and rotation of the image and the reference. This helps greatly when we are comparing real-world objects to an image because the features extracted are independent of the angle and scale of the image. Figure \ref{fig:siftres} shows an example of how features from one image could be mapped to another image even when the objects in the image have different orientations. We used OpenCV to extract SIFT descriptors for each image. The descriptors were grouped into N (N=60) clusters. Then a feature vector \(v \in \mathbb{R}^n\) was built where each direction represented a cluster and magnitude represented the count of SIFT descriptors in that cluster for the image.

\subsubsection{Background Subtraction} 
\begin{figure}
	\includegraphics[width=\linewidth]{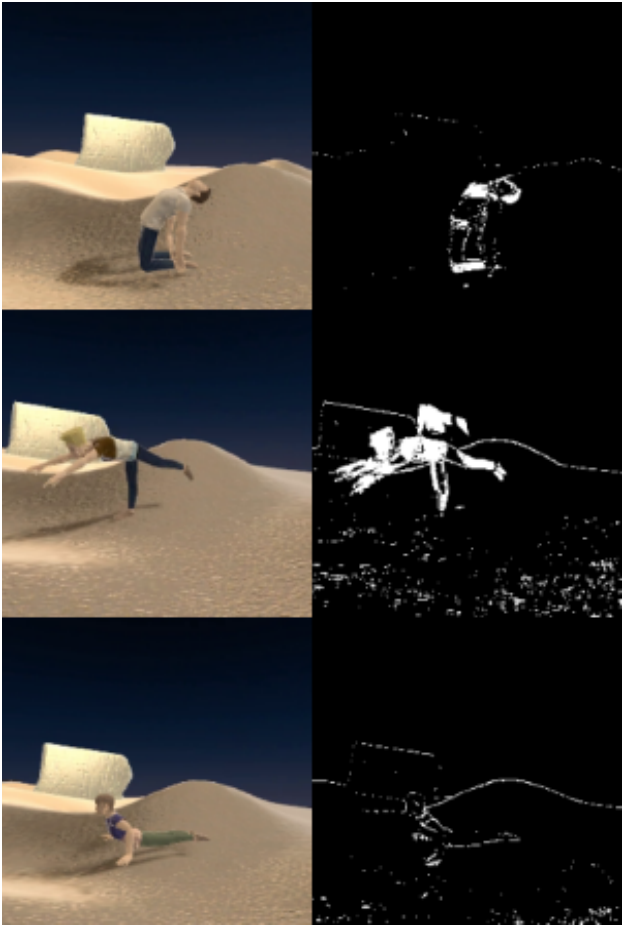}
	\caption{Image before and after background subtraction.}
	\label{fig:bgimages}
\end{figure}

Given that the medium data set and hard data set contain background, we apply a background substraction algorithm in order to test its efficacy. The algorithm we apply is the K-nearest neighbours background subtraction \cite{Zivkovic:2006qr}. We first retrieve 6 sequential frames from a video to learn and calculate the background subtracted frames, then take the last background subtracted frame as our sample data. Figure \ref{fig:bgimages} shows the results of the background subtraction filter.

\subsection{Classifier}

\subsubsection{Support Vector Machine} We used the support vector classification directly from Scikit-Learn with the default parameters: 1.0 for regularization, radial basis function as kernel and maximum iterations of 50.

\subsubsection{Logistic Regression} Again we used the logistic regression function from Scikit-Learn with default parameters: ``LBFGS'' as the solver and maximum iterations of 100.

\subsubsection{Gradient Boosting Tree} The gradient boosting tree we trained used 100 estimators, each with a maximum depth of 3 and a learning rate of 0.1.

\subsubsection{Convolutional Neural Network} Our convolutional neural network model was inspired by the VGG-16 CNN model. The model consisted of 10 convolutional layers, where the first two layers extracted 32 features and subsequent pairs of layers extracted double the number of features from the previous layers. After each pair of layers, a max pooling of size 2X2 was applied to the input features from the previous convolutional layers. Finally, two dense layers followed by a classification layer were applied to generate the model prediction.

\subsubsection{Video Vision Transformer} Finally, we apply video vision transformers to our data sets. The model we apply is the spatio-temporal attention transformer model proposed in \cite{Aarnab:2021qr}. The transformer model consists of a 3D convolutional layer, which scans through the 5 frames of size 156X156 pixels in the input data with a 2X16X16 kernel, converting the video frames into embeddings of size 128. Then 8 layers of multi-head attention each with 8 heads are applied to the embeddings to encode contextual data. Lastly, the encoded patches are transformed with softmax function to classify the input data.

\section{Evaluation}

\subsection{Baseline}

Our baseline for evaluation was the SVM model trained on the unfiltered input data scaled down to 20\% at the ``easy'' level. We felt that this model was the most natural starting point to evaluate how effective the various types of filtering and modeling techniques were and how much they contributed to the final accuracy of the models.

\subsection{Process}

We started out by training our baseline models on the ``easy'' level data set, which produced a guideline for choosing models for further training. Then combinations of filters and models with the best results were trained on the ``medium'' level data set to assess the capabilities of the models. Finally the best performance models were challenged with the ``hard'' data set for evaluation.

For every data set and model training, we splitted the sample data into three sets: training set, validation set and test set, in the ratio of 8:1:1. For classic models, validation data set was not used at all, training set was used to train and optionally validate during training depending on the model in question (GBT), and test data set was used to compute the final accuracy of the models.

\section{Results}

\begin{table*} 
	\caption{Model results for each difficulty level and model/filter combination. Action column displays results for using action only as label, A+T column displays results for using action and action type combination as label. Downsize is done by taking a 20\% scale down of the input sample data. BGSub is the background filter. }
	\centering 
	\begin{tabular}{L{0.1\linewidth} L{0.2\linewidth} C{0.05\linewidth} C{0.05\linewidth} C{0.05\linewidth} C{0.05\linewidth} C{0.05\linewidth} C{0.05\linewidth} C{0.05\linewidth} C{0.05\linewidth}}
		\toprule
		\multicolumn{2}{c}{} & \multicolumn{2}{c}{Easy} & \multicolumn{2}{c}{Medium} & \multicolumn{2}{c}{Hard}\\
		\cmidrule(r){3-8}
		Model & Filter & Action & A+T & Action & A+T & Action & A+T \\
		\midrule
		SVM & Downsize & 37\% & 29\% & 15\% & 5\% & - & - \\
		Logistic & Downsize & 43\% & 16\% & 21\% & 6\% & - & - \\
		GBT & Downsize & 71\% & 29\% & 61\% & 14\% & - & - \\
		\hline
		SVM & BGSub|Downsize & - & - & 64\% & 36\% & - & - \\
		Logistic & BGSub|Downsize & - & - & 60\% & 23\% & - & - \\
		GBT & BGSub|Downsize & - & - & 78\% & 32\% & - & - \\
		\hline
		SVM & PCA & 48\% & 42\% & 15\% & 6\% & - & - \\
		Logistic & PCA & 44\% & 17\% & 23\% & 6\% & - & - \\
		GBT & PCA & 69\% & 30\% & 24\% & 3\% & - & - \\
		\hline
		SVM & Skeleton|PCA & 48\% & 34\% & 25\% & 13\% & - & - \\
		Logistic & Skeleton|PCA & 47\% & 16\% & 35\% & 10\% & - & - \\
		GBT & Skeleton|PCA & 72\% & 25\% & 41\% & 7\% & - & - \\
		\hline
		SVM & HOG|PCA & 68\% & 54\% & 27\% & 11\% & - & - \\
		Logistic & HOG|PCA & 55\% & 21\% & 34\% & 10\% & - & - \\
		GBT & HOG|PCA & 85\% & 46\% & 30\% & 5\% & - & - \\
		\hline
		SVM & SIFT|KMean & 50\% & 43\% & 34\% & 28\% & 16\% & 9\% \\
	 	Logistic & SIFT|KMean & 70\% & 46\% & 53\% & 30\% & 33\% & 13\% \\
	 	GBT & SIFT|KMean & 71\% & 44\% & 56\% & 27\% & 32\% & 12\% \\
		\hline
		SVM & BGSub|SIFT|KMean & - & - & 33\% & 22\% & - & - \\
	 	Logistic &  BGSub|SIFT|KMean & - & - & 61\% & 33\% & - & - \\
	 	GBT &  BGSub|SIFT|KMean & - & - & 61\% & 30\% & - & - \\
		\hline
		CNN & - & - & - & 100\% & 98\% & 95\% & 92\% \\
		ViViT & - & - & - & - & - & 83\% & 38\% \\
		\bottomrule
	\end{tabular}
	\label{tab:modelres}
\end{table*}

Table \ref{tab:modelres} showed the results from training and testing combinations of 4 models and 3 filters. We started out with the easy level data set, training 3 models (SVM, Logistic, GBT) with no filters except downsizing the input images by a further 80\%. The results showed that gradient boosting tree generates the best performance out of the three models. However the results of SVM and GBT were comparable with action and action type label combinations. The models are then applied to the background subtracted ``medium'' level data set. The results show dramatic improvements over the original results on ``medium'' data set without background subtraction. What's even more surprising, is the fact that the results signficantly surpassed the results from the ``easy'' level data set as well (recall that the only difference between ``easy'' and ``medium'' level data is the inclusion of background). Suggesting that the background subtraction algorithm, in additonal to removing the background, also contributed to the filtering of extraneous information in the image data.

Three more sets of models were trained on the easy level data set with the PCA filter, the HOG plus PCA filter combination and the SIFT filter. The results showed steady improvements, where HOG plus PCA filter with SVM model achieved the best accuracy for action plus action type labels, although the SIFT filtered models performed better on average.

For the medium set, we again performed the same training and testing for all of the classic models, but this time also including the CNN model. As shown in the table, the CNN model easily outperformed all other models, achieving 98\% accuracy on the action plus action type labels. Second best performing models were the SIFT feature based models, achieving 28\% accuracy on average for action plus action type labels. Looking at the data, a general trend appeared: SIFT filtered models generally performed better. This was not surprising, because one of the key variations in our samples was the orientation of cameras, SIFT features would be able to reduce the complexity due to these variations to a great extent.

An additional set of tests were performed on the medium data set and the SIFT filter by first subtracting the background from the data (BGSub|SIFT|KMean). The results are generally better than the results without background subtraction. However, it is interesting to note that the SIFT feature extraction performed worse than the results from simply downsizing the image data after background subtraction. We contribute this outcome to possibility that the SIFT feature applied to the background subtracted data filtered out valuable information needed for more accurate classifications.

Next, we took the best performing models on medium data set and applied them to the hard data set(retraining on hard data set). The results showed that CNN model still vastly outperformed other models, the SIFT feature models performed moderately well on the action only labels, but faltered when trying to predict the action plus action type labels.

Finally, we performed one more round of predictions on the ``hard'' level data set with the video vision transformer model (ViViT). As expected, the video vision transformer model performed much better than any of the classic models. Interestingly, though, the transformer model did not outperform the CNN model. We believe this is due to the fact that the projection layer in the transformer model uses only one 3D convolutional layer, which is not enough to extract more detailed features.

\section{Model Analysis}

\subsection{Hyperparameter Search}

Although the CNN model produced superior performance by a large margin, The models with SIFT plus K-Mean clustered features did show great promise. Therefore we performed a hyper-parameter search for these models. For this exercise, we used the ``hard'' level validation data set to optimize the model parameters, Table \ref{tab:hsres} shows the results. Interesting to note is that SVM performs much better after hyper-parameter search, mainly benefiting from gamma parameter change. Logistic Regression results did not change at all, while GBT results became worse from using a much smaller set of data to estimate the parameters.

\begin{table*} 
	\caption{Best SIFT|KMean filtered model after hyperparameter search. }
	\centering 
	\begin{tabular}{L{0.1\linewidth} L{0.1\linewidth} C{0.05\linewidth} C{0.05\linewidth} R{0.5\linewidth}}
		\toprule
		\multicolumn{2}{c}{} & \multicolumn{2}{c}{Hard} & \multicolumn{1}{c}{} \\
		\cmidrule(r){3-4}
		Model & Filter & Action & A+T & Parameters \\
		\midrule
		SVM & SIFT|KMean & 40\% & 16\% & C=1 gamma=0.01 kernel=rbf \\
	 	Logistic & SIFT|KMean & 33\% & 13\% & C=0.03 penalty=l2 solver=newton-cg \\
	 	GBT & SIFT|KMean & 28\% & 10\% & learning\_rate=0.1 max\_depth=2 n\_estimators=50 \\
		\bottomrule
	\end{tabular}
	\label{tab:hsres}
\end{table*}

\subsection{T-SNE Visualization of SIFT}

\begin{figure}
	\includegraphics[width=\linewidth]{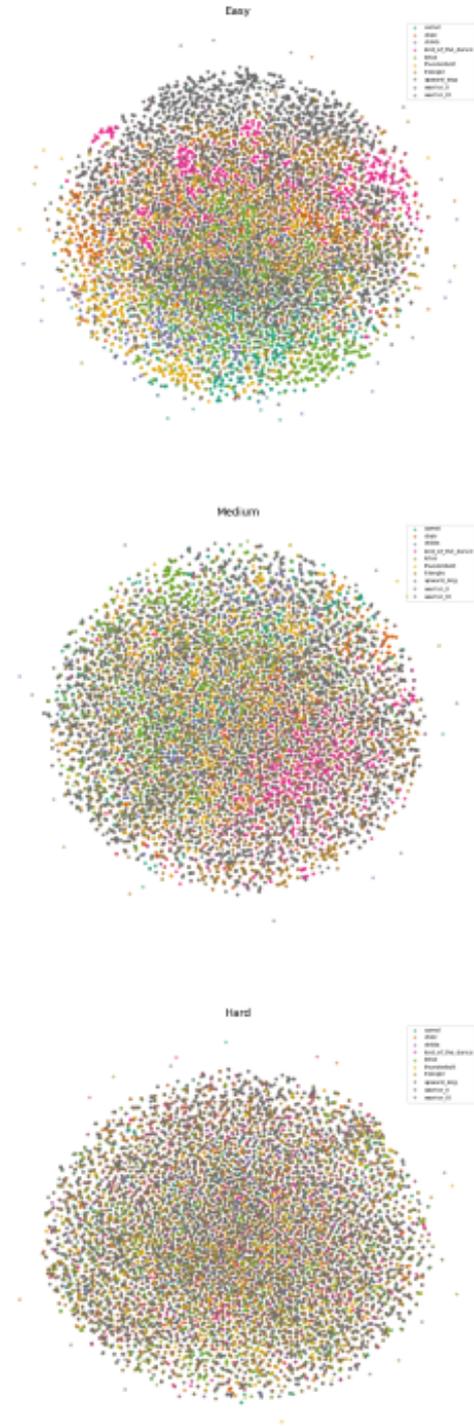}
	\caption{t-SNE visualization of SIFT plus K-Mean clustered features of ``easy'', ``medium'' and ``hard'' level data sets.}
	\label{fig:sifttsne}
\end{figure}

Next, we present a t-SNE visualization of the SIFT and K-Mean clustered features of the ``easy'', ``medium'' and ``hard'' data set in order to analyze the distinguishability of the labels within each level. Figure \ref{fig:sifttsne} displays the results. Intuitively, we would expect to see the clusters of labels to become less distinguishable as the level of difficulty increases. This is indeed the case as we move from left to right in the figures.

\subsection{PCA Analysis of HOG}

Although we did not apply PCA to the SIFT features, we did observe accuracy improvements applying PCA to the unfiltered image, HOG features and skeleton features. Therefore, we present an exploration of the effect of PCA on the extracted features. Figures \ref{fig:easypca} and \ref{fig:mediumpca} show the eigen images of the PCA decomposition of the ``easy'' and ``medium'' data set with HOG transformation. Looking at the first 16 eigen images of the easy level data set, there are no clearly discernible patterns within the images themselves, due to the orientation variations in the data set. The eign images of the medium data set, however, shows that the background of the images reflect strongly in the eigen images, reducing the accuracy of the features for action identification.

\section{Discussion}

\subsection{Error Analysis}

\begin{table} 
	\caption{Accuracy of CNN model on hard level data set.}
	\centering
	\begin{tabular}{lccc}
		\toprule
		 & Precision & Recall & F1-Score \\
		\midrule
		Macro Avg & 93\% & 93\% & 93\% \\
		Weighted Avg & 93\% & 93\% & 93\% \\
		\bottomrule
	\end{tabular}
	\label{tab:accuracycnn}
\end{table}

\subsubsection{Convolutional Neural Network} We performed a detailed analysis of the source of error in the hard level dataset based on the results we obtained from the CNN model. Table \ref{tab:accuracycnn} shows the accuracy results of the CNN model, which is very high across the board. \ref{fig:f1cnn} shows the f1-score of the model across all labels. We see that there are very few variations among them, however, the ``triangle'' action does see the lowest f1-score. Figure \ref{fig:f1cnncf} shows the confusion matrix of the ``triangle'' action labels, which clearly indicates that the confusion happens mostly within the different types of the ``triangle'' action.

Figure \ref{fig:trianglecnn} shows the incorrect predictions made by the CNN model on the ``triangle\_3'' action and action type combination. Observing these incorrect observations reveal several causes for the errors in prediction. First some angles are difficult for the model to determine what action type is performed as in image 2 of Figure \ref{fig:trianglecnn}. Some of the images simply don't contain enough of the avatar to result in an accurate prediction, such as image 7, although the model is still able to predict the action label from just an image of the leg.

\subsubsection{SVM, Logistic and GBT} Given that the accuracy of the classic models aren't very high on the hard data set, we look at the hyper-parameter searched SVM, Logistic and GBT models with SIFT K-Mean clustered features for error analysis using action plus action type label predictions on the ``hard'' level data set. Figure \ref{fig:svmconfusion}, \ref{fig:logisticconfusion} and \ref{fig:gbtconfusion} show the confusion matrix of the SVM, logistic and GBT model predictions aggregated to the action labels. We see that the ``child'' positions are the most accurate category predicted for all three models, while ``upward dog'' positions are the least accurate, which are very different from the accuracies of the CNN model.

However, considering that the models are still able to classify the samples to action labels with relatively high accuracy, it could be inferred that the models have difficulties recognizing the subtle differences in action types, which is understandable considering the low dimensionalities of the SIFT features.

Our research started out with training three different types of classic image classification models(SVM, Logistic, GBT) in combination with 3 different filtering processes(PCA, HOG, SIFT) using our novel 3-D videos data generation process. The results on the ``easy'' level data set show that the all three filtering processes significantly improve over the baseline models, demonstrating the abilities of the filters to extract salient features from the images for classifications.

Applying the models on the ``medium'' level data set shows a more subdued but still respectable model performance on the action labels, yet it is clear that the classic models have reached their limits trying to distinguish between action types. However, SIFT features are able to help the models achieve higher performance through extracting features that are independent of the background, showcasing its advantages.

Finally, the ``hard'' level data set results in uniformly poor performance across the classic models, due to increased level of camera orientation difference and additional dynamic background, testing the limits of the SIFT features. Yet the CNN deep learning model is still able to command above 90\% accuracy, unequivocally demonstrating the capabilities of the new technique.

Through this process we demonstrated how a flexible and comprehensive 3-D data generation process could tremendously improve our research capability and quality, opening doors to many future research topics.

\section{Future}

Our work is far from finished. The research results so far are only a starting point, produced in order to showcase what could be possible with the new data generation process. Many topics remain unexplored, some of the immediate next research areas include applying 3D convolutional neural networks on the colored image data, incorporation of optical flow information and applying more complex 3D convolutional neural network encoder with transformer models for sequential images embedding and action classification, for both single and sequential actions.

In Additon, with enhancements to the data simulation engine, we would be able to explore more complex actions and blending of actions. Multiple objects in the scene would allow us to evaluate localized multiple action recognition \cite{Wu:2023qr}. Eventually, we could develop a system with the capability to recognize continuous actions for multiple objects within a video data stream.

\phantomsection
\bibliographystyle{unsrt}
\bibliography{t0004.bib}
\clearpage
\begin{appendices}
\onecolumn
\section*{APPENDIX}
\subsection{Action Figures}

\begin{figure}[H]
	\includegraphics[width=\linewidth]{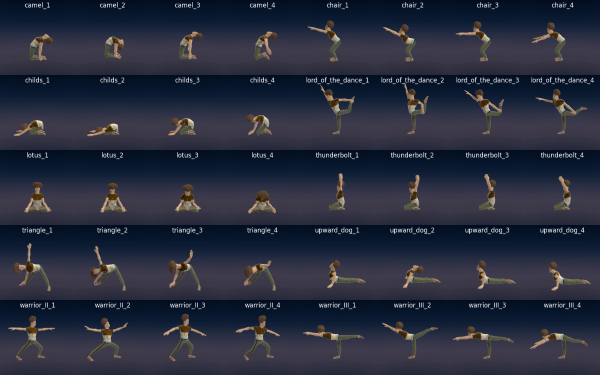}
	\caption{10 yoga poses, each with 4 different variants.}
	\label{fig:actionsfig}
\end{figure}

\subsection{Model Errors and Results}

\begin{figure}[H]
	\includegraphics[width=\linewidth]{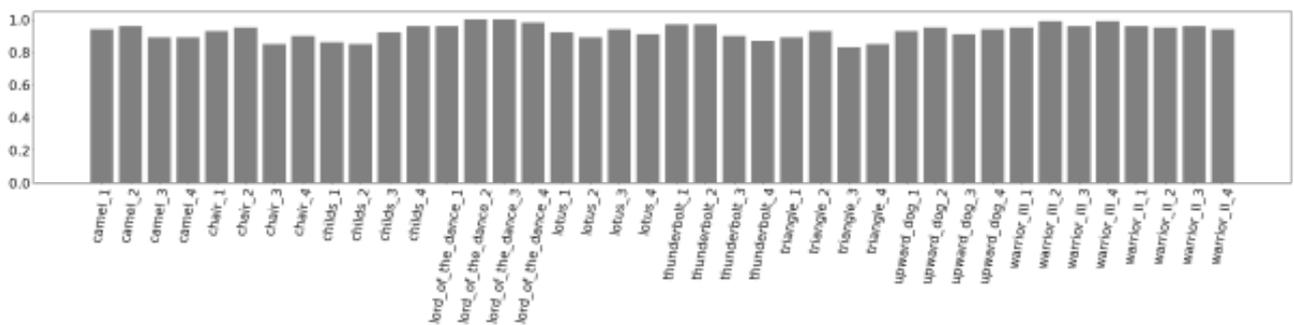}
	\caption{F1 scores of the CNN model across labels.}
	\label{fig:f1cnn}
\end{figure}

\begin{figure}[H]
	\includegraphics[width=\linewidth]{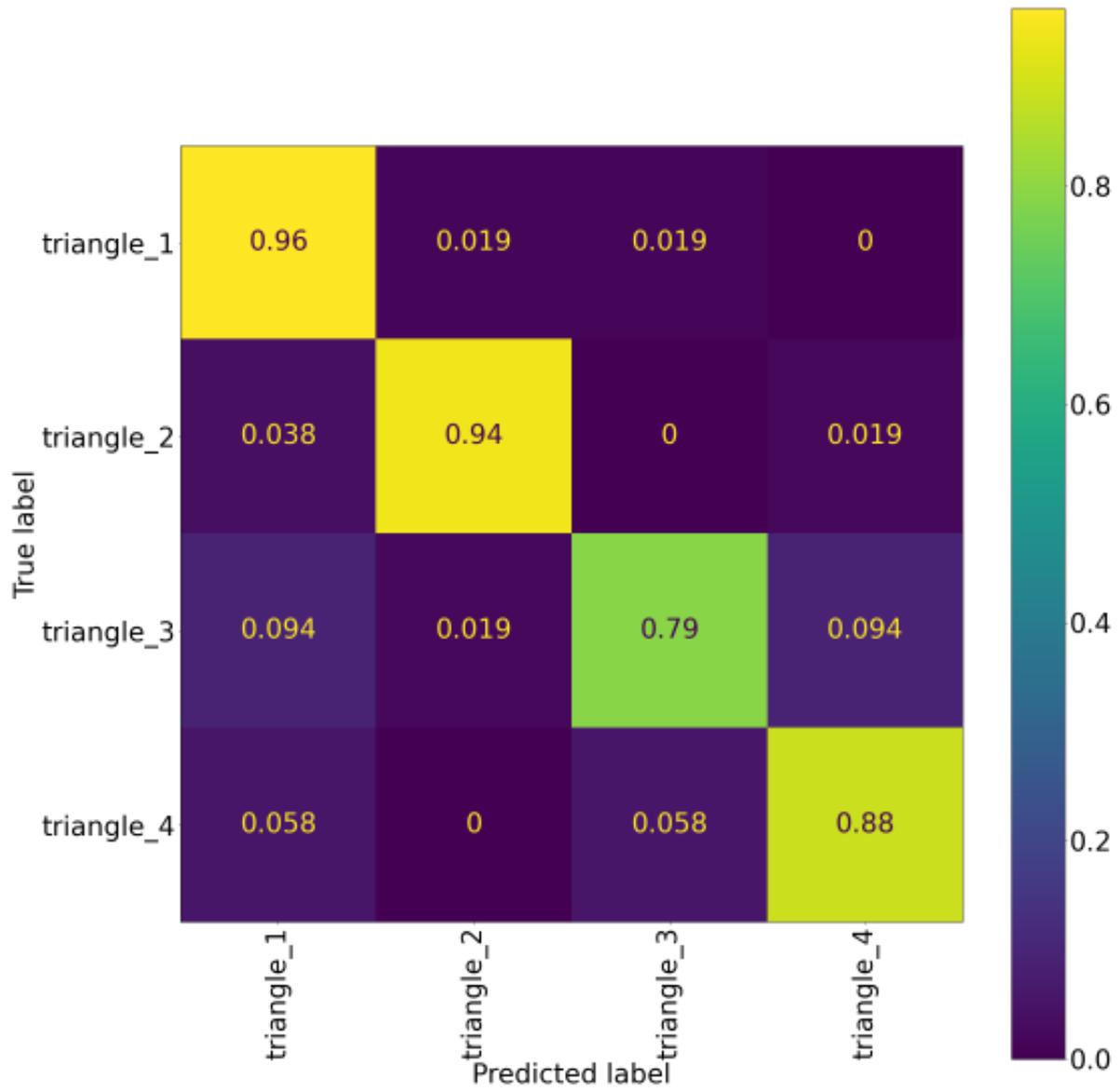}
	\caption{CNN model confusion across ``triangle'' action labels.}
	\label{fig:f1cnncf}
\end{figure}

\begin{figure}[H]
	\includegraphics[width=\linewidth]{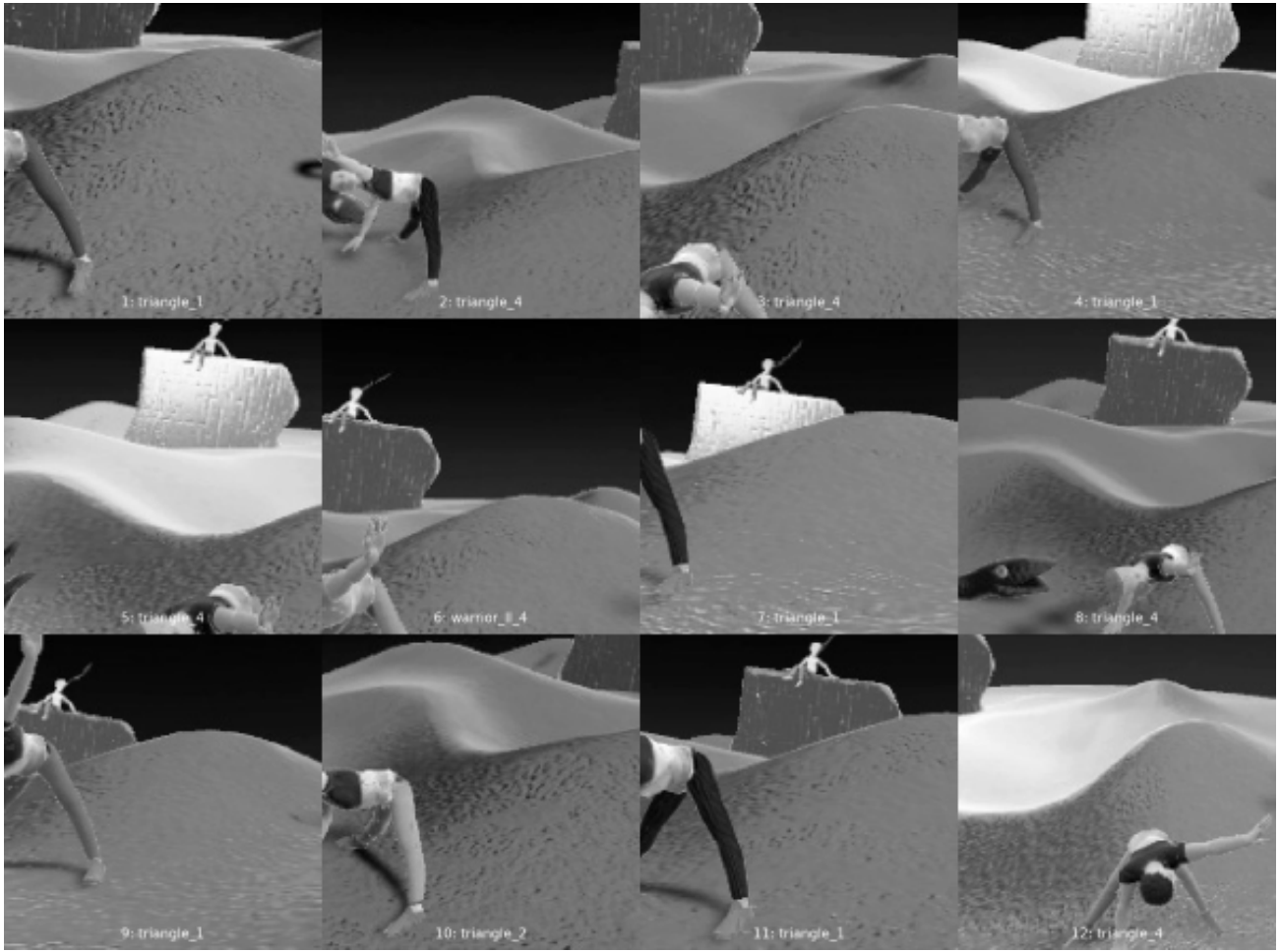}
	\caption{CNN model confusion across ``triangle'' action labels.}
	\label{fig:trianglecnn}
\end{figure}

\begin{figure}[H]
	\includegraphics[width=\linewidth]{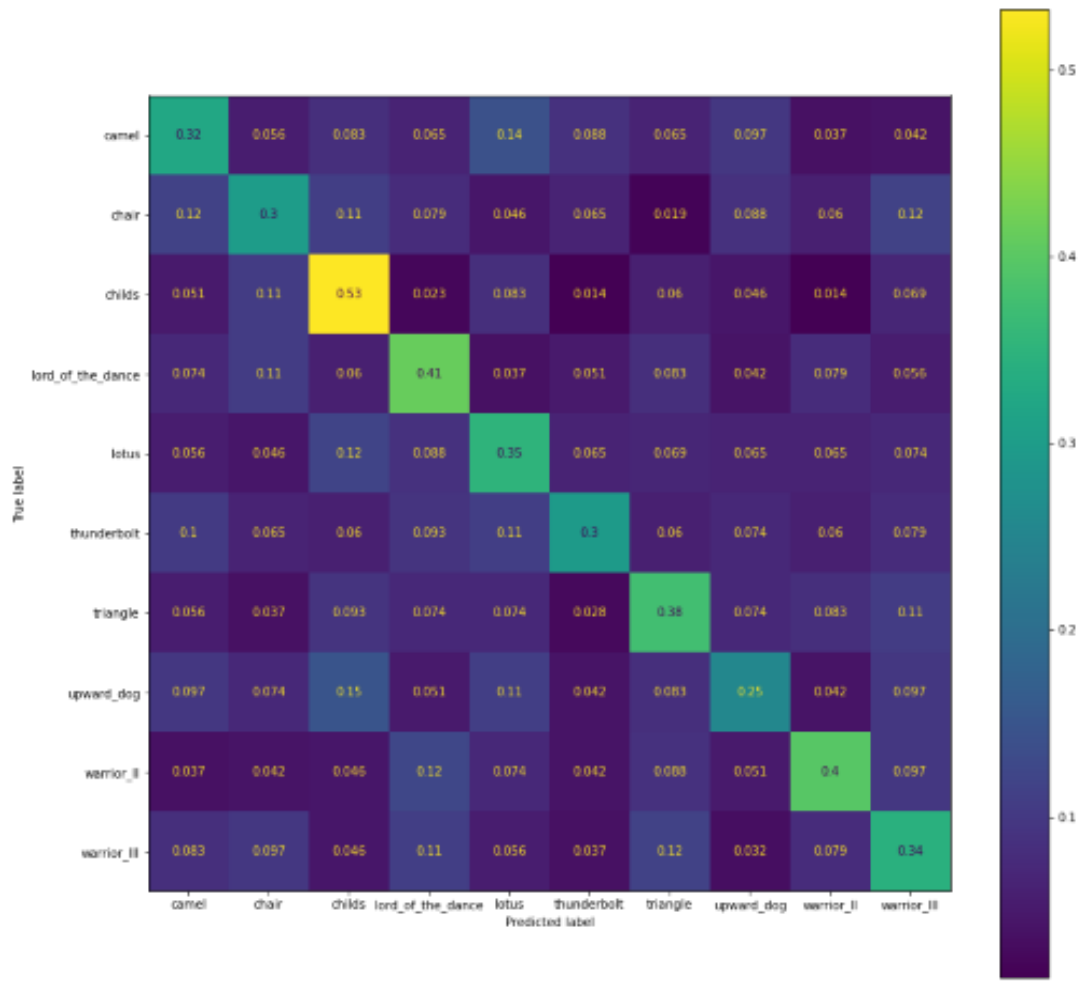}
	\caption{SVM model confusion matrix aggregated to the action labels.}
	\label{fig:svmconfusion}
\end{figure}[H]

\begin{figure}[H]
	\includegraphics[width=\linewidth]{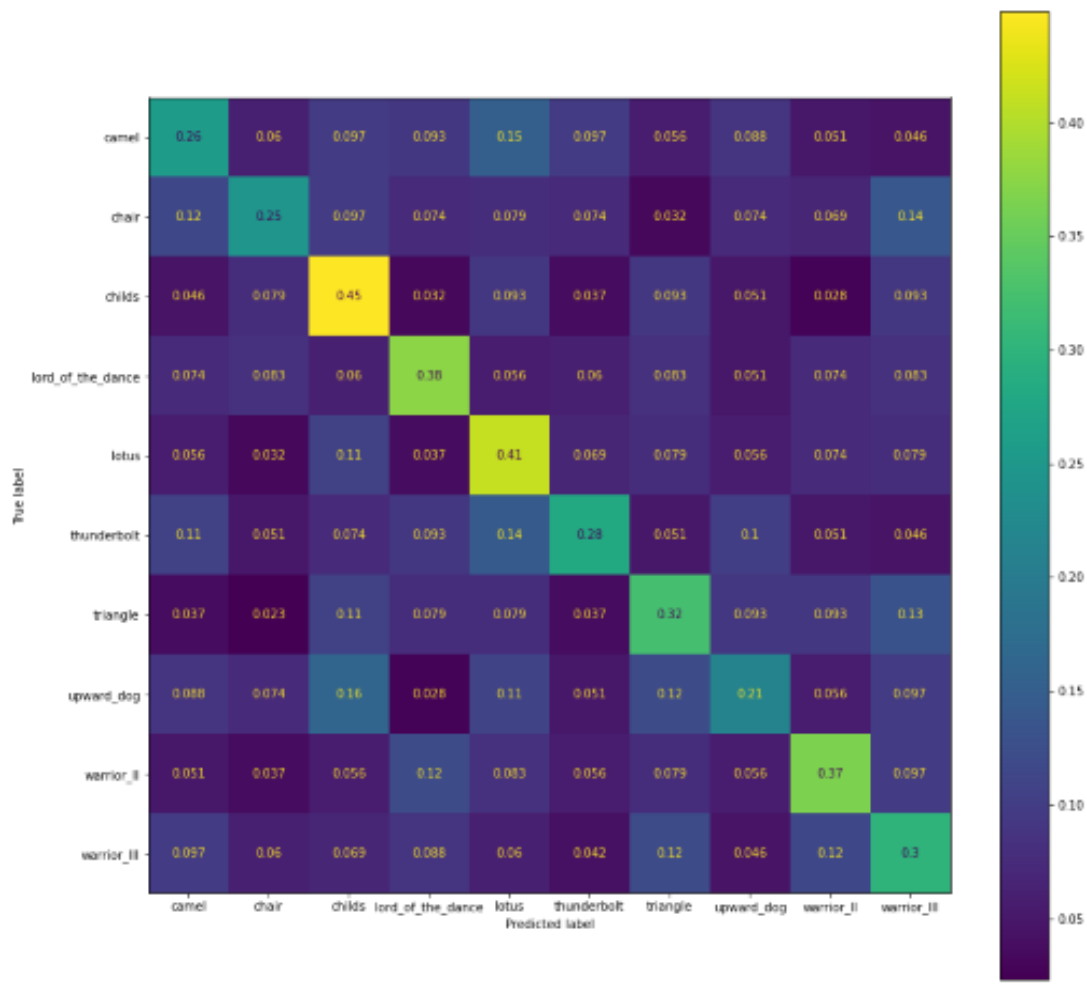}
	\caption{Logistic model confusion matrix aggregated to the action labels.}
	\label{fig:logisticconfusion}
\end{figure}

\begin{figure}[H]
	\includegraphics[width=\linewidth]{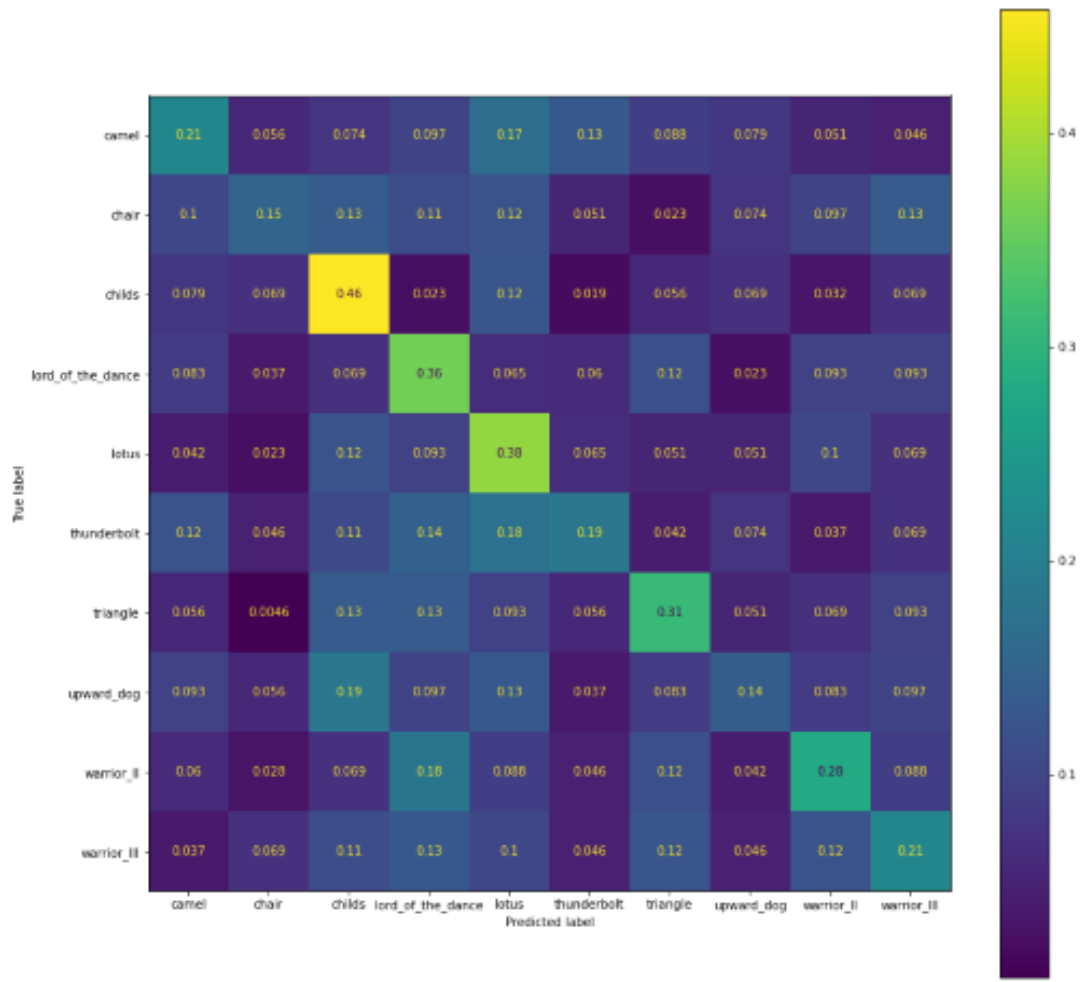}
	\caption{GBT model confusion matrix aggregated to the action labels.}
	\label{fig:gbtconfusion}
\end{figure}

\subsection{PCA Decomposition of HOG Features}

\begin{figure}[H]
	\includegraphics[width=\linewidth]{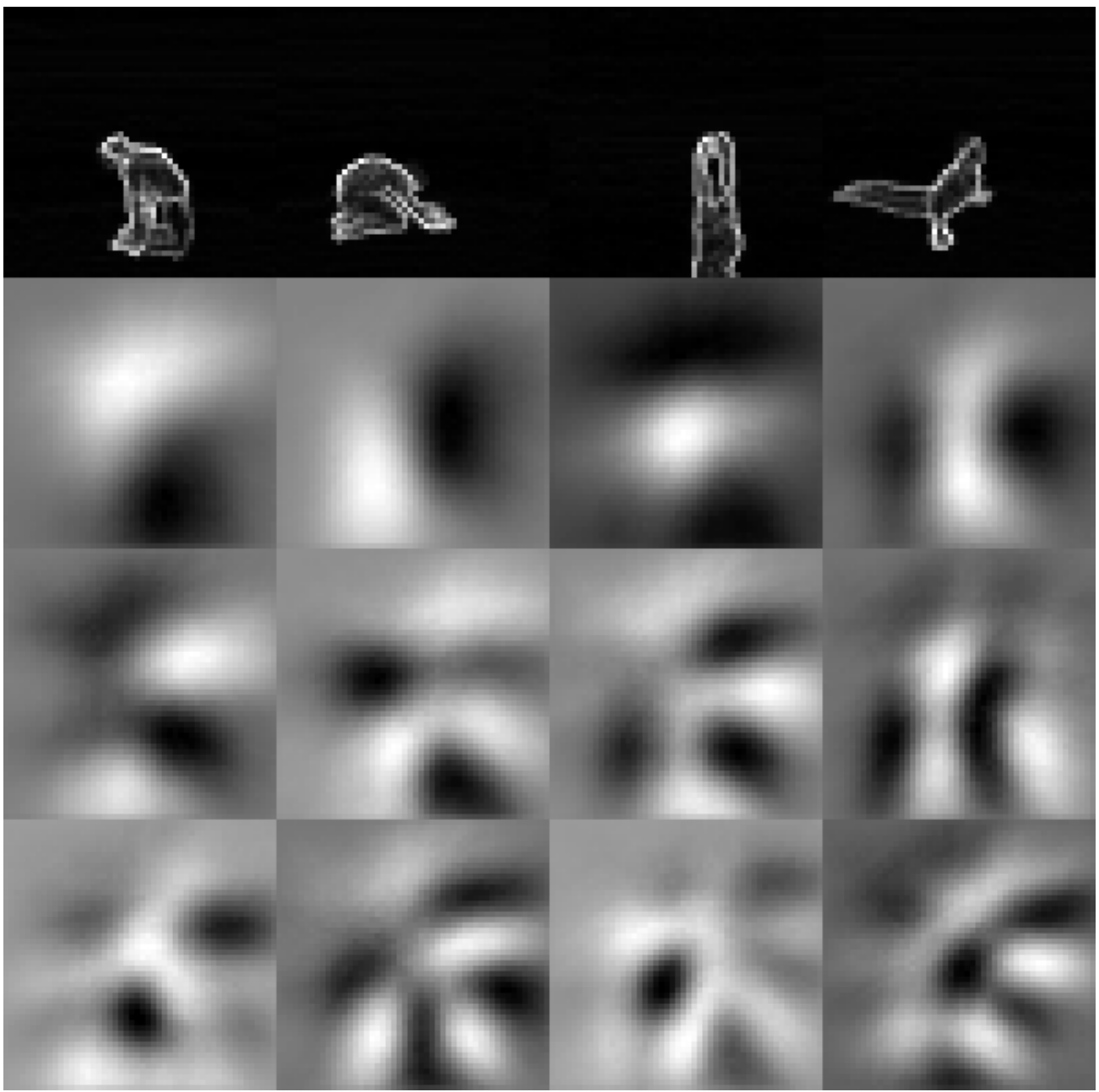}
	\caption{PCA decomposition of easy level data set with HOG filter. First row contains sample HOG images from the data set.}
	\label{fig:easypca}
\end{figure}

\begin{figure}[H]
	\includegraphics[width=\linewidth]{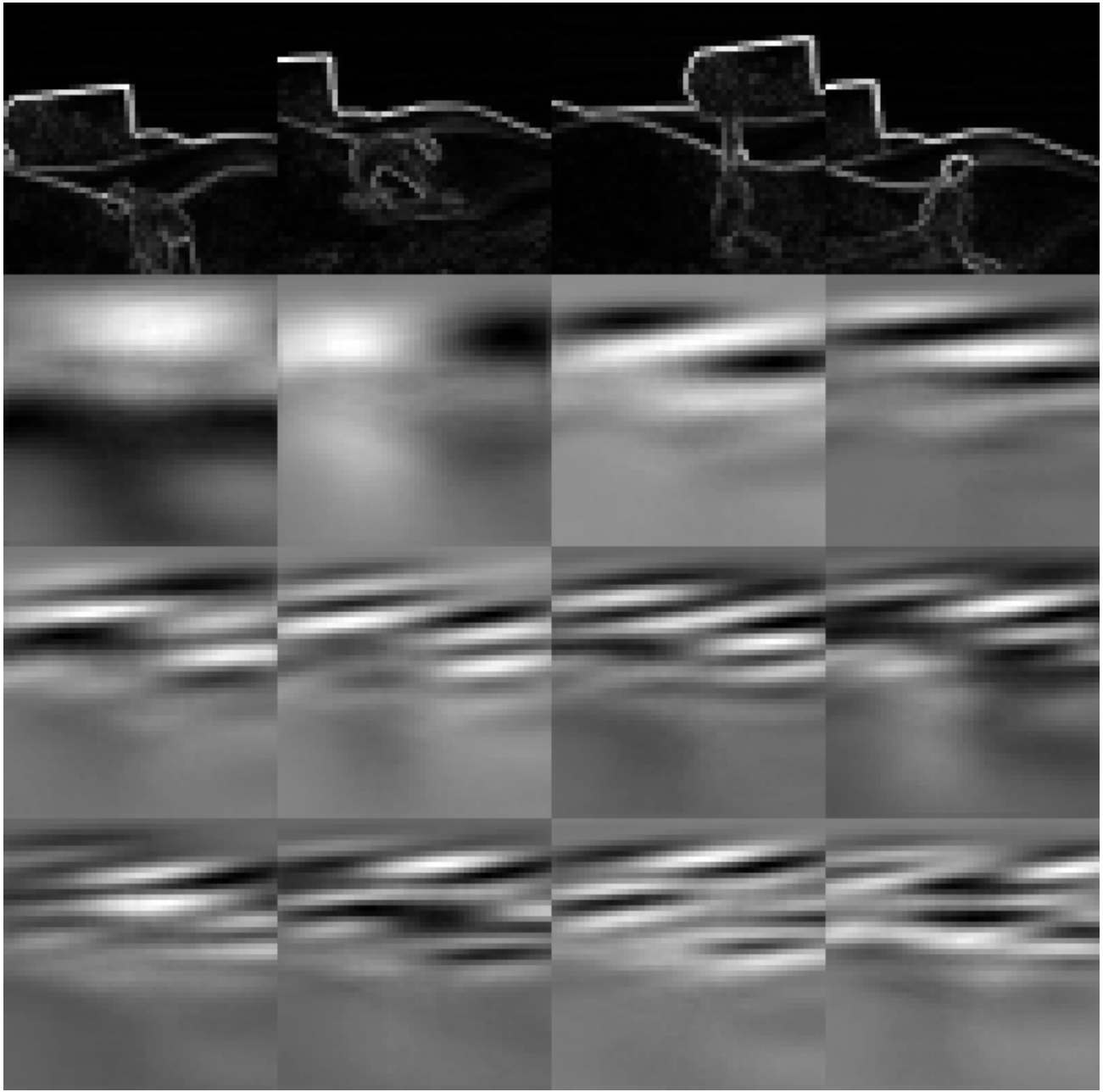}
	\caption{PCA decomposition of medium level data set with HOG filter. First row contains sample HOG images from the data set.}
	\label{fig:mediumpca}
\end{figure}

\end{appendices}
\end{document}